\documentclass[10pt,conference,a4paper,compsoc]{IEEEtran}
\usepackage{graphicx,times}
\usepackage{amsmath,amssymb}

\usepackage{enumitem,kantlipsum}
\usepackage[font=small,labelfont=bf]{caption}

\usepackage[left=1.31cm,top=1.89cm,right=1.31cm,bottom=4.3cm]{geometry}

\begin{document}

\title{{\vspace{0.26in}
\bf \Large Epipolar Geometry Based On Line Similarity} }

\author{ 
Gil Ben-Artzi ~~~~~ Tavi Halperin ~~~~~ Michael Werman ~~~~~ Shmuel Peleg\\
\emph{School of Computer Science and Engineering} \\ 
\emph{The Hebrew University of Jerusalem, Israel}\\}

\date{}

\maketitle

\section*{\centerline{Abstract}}
\textnormal{\textit{
It is known that epipolar geometry can be computed from three epipolar line correspondences but this computation is rarely used in practice since there are no simple methods to find corresponding lines. Instead, methods for finding corresponding points are widely used. This paper proposes a similarity measure between lines that  indicates whether two lines are corresponding epipolar lines and enables finding epipolar line correspondences as needed for the computation of epipolar geometry.
}}

\textnormal{\textit{
A similarity measure between two lines, suitable for video sequences of a dynamic scene, has been previously described. This paper suggests a stereo matching similarity measure suitable for images. It is based on the quality of stereo matching between the two lines, as corresponding epipolar lines yield a good stereo correspondence.
Instead of an exhaustive search over all possible pairs of lines, the search space is substantially reduced when two corresponding point pairs are given. 
}}

\textnormal{\textit{
We validate the proposed method using real-world images and compare it to state-of-the-art methods. We found this method to be more  accurate  by a factor of five compared to the standard method using seven corresponding points and  comparable to the 8-point algorithm.
}}

\section{Introduction}

The fundamental matrix is a basic building block of multiple view geometry and its computation is the first step in many vision tasks. The computation  is usually based  on pairs of corresponding points. Matching points across images is error prone and many subsets of points need to be sampled until a good solution is found. In this work we address the problem of robustly estimating the fundamental matrix from line correspondences, given a similarity measure between lines.

The best-known algorithm, adapted for the case of fundamental matrix, is the eight point algorithm by Longuet-Higgins \cite{longuet1981computer}. It was made practical by Hartley \cite{hartley1997defense}. The overall method is based on normalization of the data, solving a set of linear equations and enforcing the rank 2 constraint \cite{luong1996fundamental}.  The requirement of {\it eight} point correspondences can be relaxed to seven. This results in  a cubic  equation with  one or three real solutions. The estimation from  7 points is very sensitive to  noise. These methods are often followed by a  non-linear optimization step. 

The fundamental matrix can also be computed from three matching epipolar lines \cite{hartley2003multiple}. Given three such correspondences, the one dimensional homography between the lines can be recovered as the epipolar lines in each of the images intersect at the epipoles. The 3 degrees of freedom for the 1D homography together with the 4 degrees of freedom of the epipoles yield the required 7 degrees of freedom needed to compute the fundamental matrix.  A method for computation of the fundamental matrix by correspondences of only two SIFTs descriptors is presented in \cite{goshen2008balanced}. They used the fact that each feature point is accompanied by its local descriptor. They proposed three additional correspondences which lie within the descriptor window. This method is less accurate than the 7 and 8 points methods which are used as our baseline.

There are a few papers using corresponding epipolar lines to compute epipolar geometry, but these are only applicable to videos of dynamic scenes \cite{sinha2010camera,Calibration2016CVPR,Calibration2016ECCV}. The most relevant paper is \cite{Calibration2016ECCV}, describing a similarity between lines based on scene dynamics.

We present a new similarity measure between lines, and utilize it to robustly compute the fundamental matrix. This similarity measure is based on the brightness consistency (stereo matching) that exists between corresponding epipolar lines. We can reduce the search space for corresponding epipolar lines by giving two corresponding points. This is compared to the 7 or 8 points normally required. 

Fundamental matrix computation from points correspondences has to take into account mistaken correspondences (outliers). Using a RANSAC approach, multiple subsets of points are sampled, so that with high probability one subset will not include any outlier. Using only 2 points as needed in our method substantially reduces  the number of samples. For example, if 50\% of correspondences are correct, and we require a probability of 99\% that one subset will have no outliers, we will need to select $1,177$ 8-point subsets, $588$ 7-point subsets, and only $17$ subsets of 2-points.

\section{Previous Work}

\subsection{Computing the Fundamental Matrix}

The fundamental matrix is a $3 \!\times\! 3$ homogeneous rank two matrix with  seven degrees of freedom. There are various  formulations that have been considered  to produce a minimal parameterization with only seven parameters \cite{hartley2003multiple}.  

The most common parameterization  is from the correspondences of seven points and can be computed as the null space of a $7 \!\times\! 9$ matrix. The rank two constraint leads to a cubic equation with one or three possible
solutions.

The method we will follow is based directly on the epipolar geometry entities. The fundamental matrix is represented by the epipoles and the epipolar line homography. Each of the two epipoles accounts for two parameters. The epipolar line homography represents the 1D-line homography between the epipolar pencils and accounts for three degrees of freedom. 

\subsection{Finding Corresponding Epipolar Lines}

Previous methods found corresponding epipolar lines from videos of dynamic scenes. Sinha and Pollefeys \cite{sinha2010camera} used silhouettes to find corresponding epipolar lines for calibration of a network of cameras, assuming a single moving silhouette in a video. Ben-Artzi et al. \cite{Calibration2016CVPR} accelerated Sinha's method using a similarity measure for epipolar lines. The similarity measure is a generalization of Motion Barcodes \cite{ben2015event} to lines.  This line motion barcode was also used in \cite{Calibration2016ECCV} to find corresponding epipolar lines.

\subsection{Motion Barcodes of Lines}
\label{Sec:Barcodes}

Motion Barcodes of Lines were used in the case of synchronized stationary cameras viewing a scene with  moving objects. Following  background subtraction  \cite{cucchiara2003detecting} we obtain a binary video, where "0"  represents static background and "1"   moving objects.

\begin{figure}
\centering
\includegraphics[width=1\linewidth]{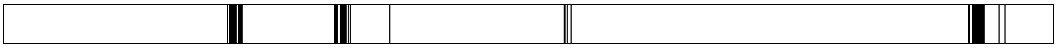}			
\caption{
A motion barcode $b$ of a line $l$ is a vector in $\{0,1\}^N$.  The value of $b_l(i)$ is "1" when a moving object intersects the line
in frame $i$ ({\it black entries}) and "0" otherwise ({\it white entries}).
\label{fig:point_barcode}}
\end{figure}

Given such a video of $N$ binary frames, the Motion Barcode of a given image line $l$ \cite{Calibration2016CVPR} is a binary vector $b_l$ in $\{0,1\}^N$. 
$b_l(i)=1$ iff a silhouette of a foreground object intersects at least one pixel of line $l$ at the $i^{th}$ frame. An example of a Motion Barcode is shown in Fig~\ref{fig:point_barcode}.  

\begin{figure}
\centering
\includegraphics[width=0.85\linewidth]{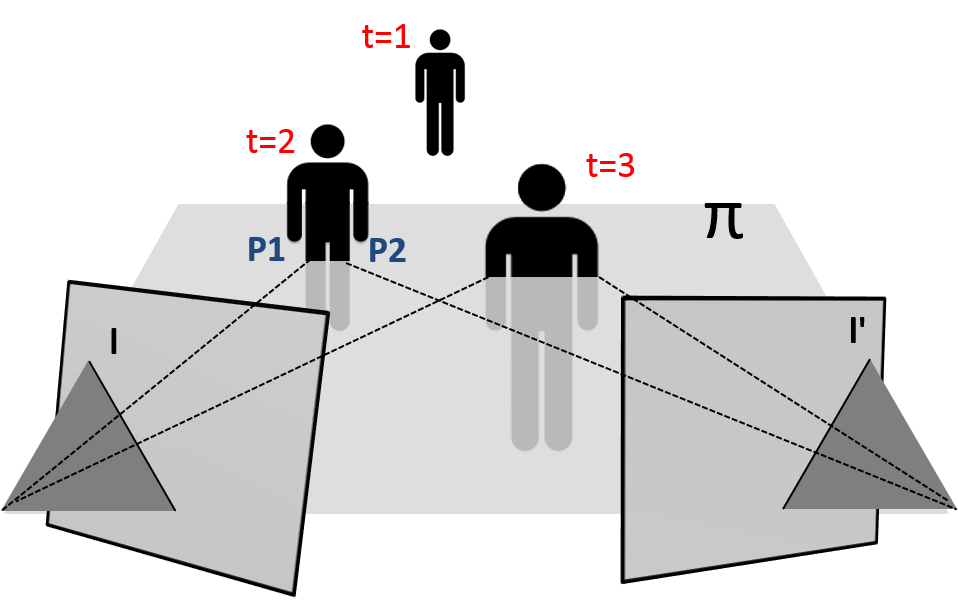}
\caption{Illustration of a scene with a moving object viewed by two video cameras. The lines $l$ and $l'$ are corresponding epipolar lines, and $\pi$ is the 3D epipolar plane that projects to $l$ and $l'$. At time $t=1$ the object does not intersect the plane $\pi$, and thus does not intersect $l$ or $l'$ in the video.  At times $t=2,3$ the object  intersects the plane $\pi$, so the projections of this object on the cameras does intersect the epipolar lines $l$ and $l'$. The motion barcodes of both $l$ and $l'$ is $(0,1,1)$ }
\label{fig:line_barcode}
\end{figure}

The case of a moving object seen by two cameras is illustrated in Fig.~\ref{fig:line_barcode}. If the object intersects the epipolar plane $\pi$ at frame $i$, and does not intersect the plane $\pi$ at frame $j$, both Motion Barcodes of lines $l$ and $l'$ will be $1,0$ at frames $i,j$ respectively. Corresponding epipolar lines therefore have highly correlated Motion Barcodes.

\subsubsection*{Similarity Score Between Two Motion Barcodes}

It was suggested in \cite{ben2015event} that a good similarity measure between motion barcodes $b$ and $b'$ is their normalized cross correlation.

\subsection{Stereo Matching}

Depth from two stereo images is traditionally computed by matching along corresponding epipolar lines. Our hypothesis is that stereo matching will be more successful when applied to corresponding epipolar lines, rather than to random, unrelated lines. The success of stereo matching along two lines is our indicator whether these two lines are corresponding epipolar lines.

Many different stereo matching methods exist (see Scharstein and Szeliski \cite{scharstein2002taxonomy} for a  survey). The stereo matching methods can be roughly divided to global and local methods. Since we are not interested in estimating an accurate per-pixel disparity, but only in line-to-line matching, we used a dynamic programming stereo method. Dynamic programming is the simplest and fastest global stereo algorithm, is relatively robust, and it gives the optimal solution for scanline matching.

\section{The Stereo Matching Similarity}
\label{sec:stereo}

\begin{figure}[tb]
\centering
a) \includegraphics[width=0.43\linewidth]{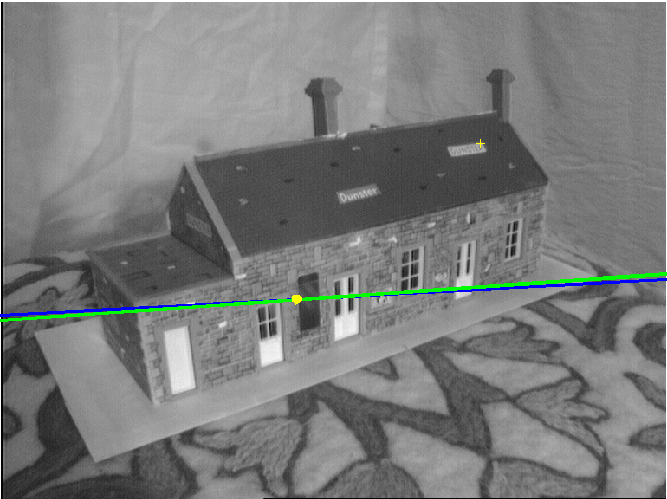}~
b) \includegraphics[width=0.43\linewidth]{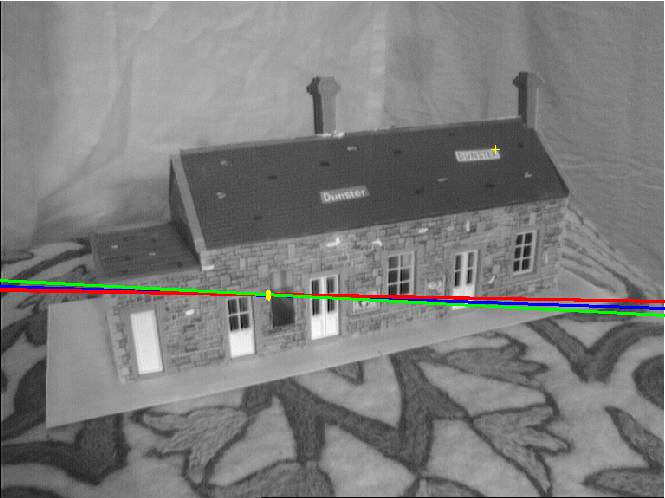}\\~\\
c) \includegraphics[height=2.4cm, width=0.43\linewidth]{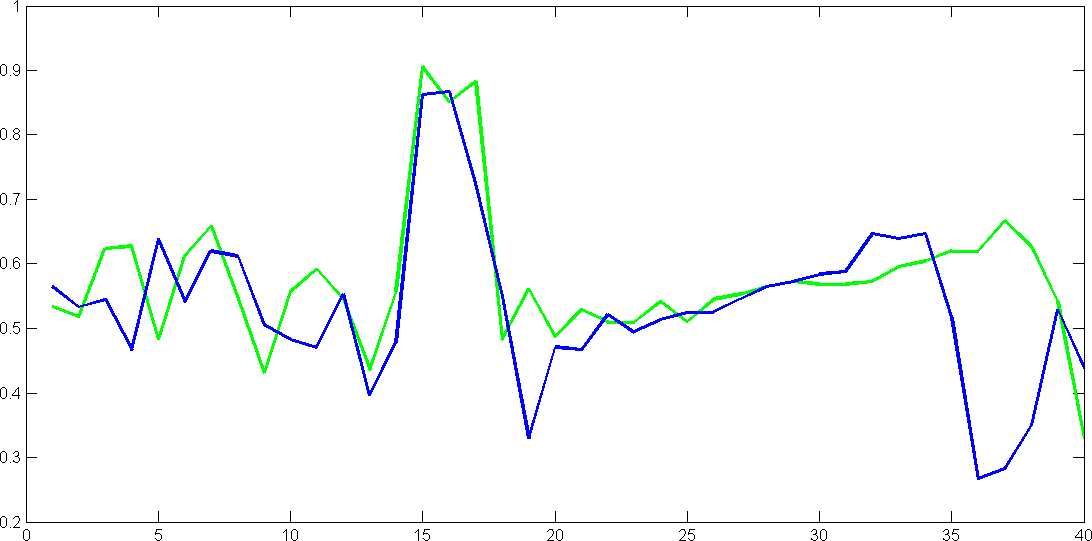}~
d) \includegraphics[height=2.4cm, width=0.43\linewidth]{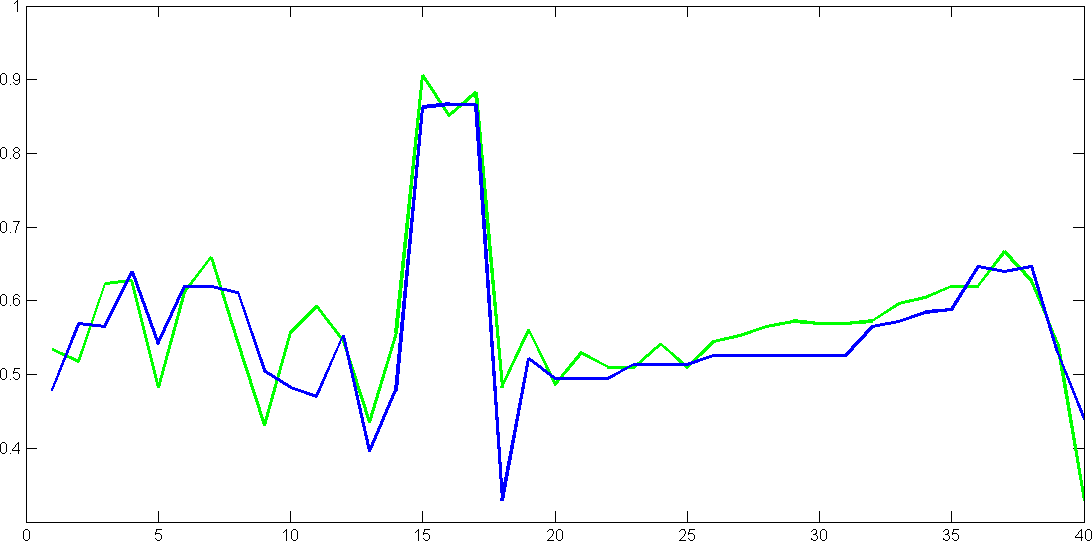}

\caption{Matching epipolar lines using stereo matching. (a-b) Matching epipolar lines across two view. Computed epipolar lines by our method (green) are very close to the ground truth (blue). 
(c) The intensity profile of the two corresponding lines found by our approach. (d) Intensity profile after warping due to stereo correspondence.}
\label{fig:lines}
\end{figure}

Two corresponding epipolar lines are projections of the scene intersected by the same epipolar plane. We assume that the intensities along corresponding epipolar lines are related through stereo disparities, as is traditionally used in stereo. Stereo depth is computed by matching points along corresponding epipolar lines. 
The stereo matching similarity between two lines, $l_1, l_2$, is therefore related to their stereo matching score as defined in Eq.~\ref{eq:linesScore}.

As the compared lines are usually not aligned with the images axes, they should be re-quantized to equidistant points along them.
Let $x_i$ be the 2D coordinates of equidistant points along line $l_1$, and let $y_i$ be the 2D coordinates of equidistant points along line $l_2$. The similarity between the two lines is based on their intensity differences.
It is formulated by the well known stereo matching equation \cite{scharstein2002taxonomy,szeliski2006comparative}
given the two lines $l_1$ and $l_2$, and the disparity $d_i$ for every point $x_i$ on $l_1$:

\begin{equation}
C(d;l_1,l_2)=\sum_{i=1}^n \phi(d_i;r) + 
\sum_{i=2}^{n} \psi (d_i;\alpha,\lambda),
\label{eq:linesScore}
\end{equation}
where $\phi(d_i;r)$ is the truncated $L_2$  intensity difference:
$$
\phi(d_i;r)=min\{( I_1(x_i)-I_2(y_{i+d_i}) )^2,r\}
$$
and $r=50^2$ \cite{scharstein2002taxonomy}. The smoothness term $\psi$ of the disparities $d_i$ is given by:
$$
\psi (d_{i};\alpha,\lambda)=min(\lambda \cdot (d_i-d_{i-1})^2, \alpha).
$$
where in our implementation we selected $\lambda=2$ and $\alpha=3$. The distance between two lines is the minimal disparity,  $C^{*}$:
\begin{equation}
C^{*}= \min_{d \in \mathbb{Z}^n } \big\{C(d;l_1,l_2) \big\},
\label{eq:disparity}
\end{equation}

Since we find the minimal disparities in Eq.~\ref{eq:disparity} using dynamic programming, the order constraint commonly used in stereo matching, $d_{i+1} \ge d_{i}$, is naturally obtained.

\section{Fundamental Matrix from Corresponding Lines}	
\label{section:ransac}

Given candidate corresponding pairs of epipolar lines between cameras $A$ and $B$, our goal is to find the fundamental matrix $F$ between the cameras. This will be carried out using a RANSAC approach, as not all of our correspondence candidates are correct. Among all possible pairs of lines, one in each image, we pick as candidate pairs those pairs having highest stereo matching similarity. To overcome the wrong correspondences among those candidates pairs, we use RANSAC \cite{Fischler-Ransac:1981}.

In each RANSAC trial, two pairs of candidate corresponding epipolar lines are selected. This gives two candidates for epipolar lines in each camera, and the epipole candidate for this camera as the intersection of these two epipolar lines. Next, an additional pair of corresponding epipolar lines is found from lines incident to these epipoles. The homography $H$ between corresponding epipolar lines is computed from these three pairs of epipolar lines, described in detail in Sec.~\ref{section:Hcompute}.

The consistency score of a proposed homography $H$ depends on the number of inliers that $H$ transforms successfully as described in Section~\ref{section:Hconsistency}. Given the homography $H$, and the epipole $e'$ in $B$, the fundamental matrix $F$ is \cite{hartley2003multiple}: 
\begin{align}\label{equation:fFromH}F=[e']_x H \end{align}

\subsection{Computing the Epipolar Line Homography}
\label{section:Hcompute}

We compute the Epipolar Line Homography using  RANSAC. We sample pairs of corresponding epipolar line candidates with a probability proportional to their stereo matching similarity as in Eq.~\ref{eq:disparity}.
Given 2 sampled pairs $(l_1, l_1')$ and $(l_2, l_2')$, corresponding epipole candidates are: $e = l_1 \times l_2$ in Camera A, and $e' = l_1' \times l_2'$ in Camera B. 
Given $e$ and $e'$, a third pair of corresponding epipolar line candidates, $(l_3,l_3')$, is selected such that they pass through the corresponding epipoles.

The homography $H$ between the epipolar pencils is calculated by the homography DLT algorithm  \cite{hartley2003multiple}, using the 3 proposed pairs of corresponding epipolar lines.

\subsection{Consistency of Proposed Homography}
\label{section:Hconsistency}

Given the homography $H$, a consistency measure with all epipolar line candidates is calculated. This is done for each corresponding candidate pair $(l,l')$ by comparing the similarity between $l'$ and $\tilde{l'}= H l$. A perfect consistency should give $l' \cong \tilde{l'}$.

Each candidate line $l$ in  $A$ is transformed to  $B$ using the homography $H$ giving $\tilde{l'}= H l$. To measure the similarity in  $B$ between $l'$ and $\tilde{l'}$ we use the image area enclosed between the lines.

The candidate pair $(l, l')$ is considered an inlier relative to the homography $H$ if the area between  $l'$ and $\tilde{l'}$ is smaller than a predefined threshold. In the experiments in Sec.~\ref{sec:Experiments} this threshold was taken to be 3 pixels times the width of the image. The consistency score of $H$ is the number of inliers among all candidate lines.

\section{Fundamental Matrix From Two Points}

The process described in Sec~\ref{section:ransac} starts by computing the stereo matching similarity between all possible pairs of lines in the two images. This is computationally expensive, and many wrong correspondences will still have a good score. In order to reduce the computation time and reduce the number of mismatches, we assume that two pairs of corresponding points are given. Note that it is still much lower than the 7 or 8 pairs of points usually needed.

\subsection{Problem Formulation}

Given two images $I_1,I_2$ with two pairs of corresponding points $(p_1, p_2)$ and $(q_1, q_2)$, we want to estimate the fundamental matrix $F$ between the images.
 
We start by seeking epipolar lines for each pair of corresponding points, using the consistency of the intensities along the lines. This consistency is computed from an optimal stereo matching along these lines. Each corresponding pair of points gives us an epipolar line in each images, and two pairs of corresponding points give us two epipolar lines in each image. Once we find the  epipolar lines for the two pairs of corresponding points, the intersections in each image of the two epipolar lines gives the epipole. The third epipolar line needed for computing $F$ is found from lines passing through the recovered epipoles. 

In later subsections we introduce an iterative approach to compute the fundamental matrix using a RANSAC based algorithm for finding epipolar lines and epipoles.

\begin{figure}[tb]
\centering
\includegraphics[width=0.6\linewidth]{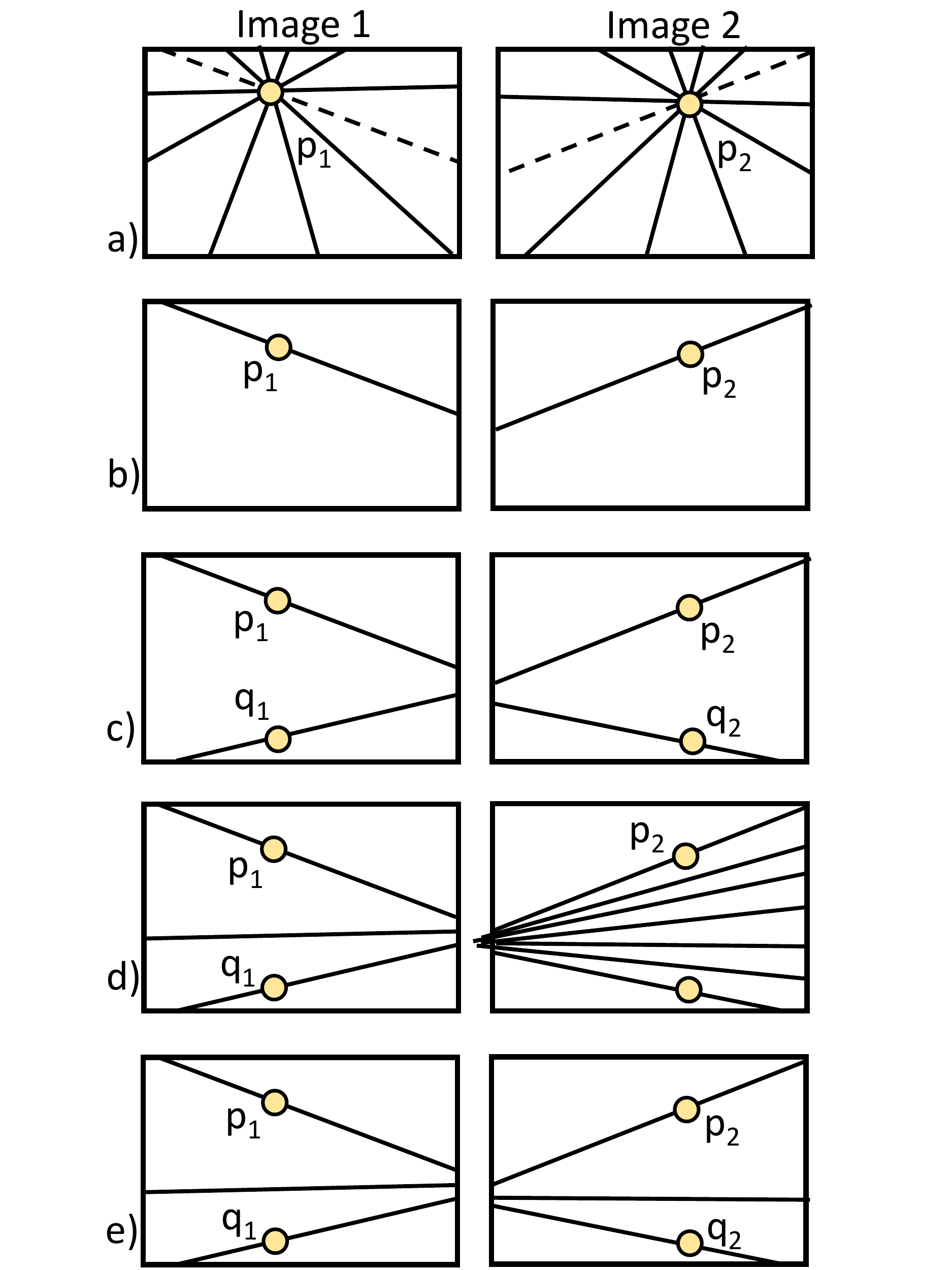}
\caption{The Two-Points algorithm.
(a) Given a pair of corresponding points, $p_1$ in Image 1 and $p_2$ in Image 2, we examine a pencil of lines passing through each point.
(b) Each line from the pencil  $p_1$,  together with the line from the pencil of $p_2$, having the best stereo matching score, is selected as a possible corresponding pair of epipolar lines.
(c) A pair of corresponding epipolar lines, passing through the second pairs of corresponding points, $q_1$ and $q_2$, is selected in the same manner. Their intersection in each image gives the third point - the epipole (which may be outside the image).
(d) A bisector epipolar line is selected in Image 1, and a corresponding epipolar lines is searched for in the pencil of lines passing through to the epipole in Image 2.
(e) The last stage of the process gives us three corresponding epipolar lines, from which the epipolar line homography is computed.}
\label{fig:overview}
\end{figure}

\subsection{Computation of Epipolar Line Homography}

Following are the steps for the computation of the epipolar line homography between two images, when two corresponding pairs of points in two images, $(p_1, p_2)$ and $(q_1, q_2)$, are given. The process is outlined in Fig.~\ref{fig:overview}.

\begin{figure}[tb]
\centering
a) \includegraphics[width=0.43\linewidth]{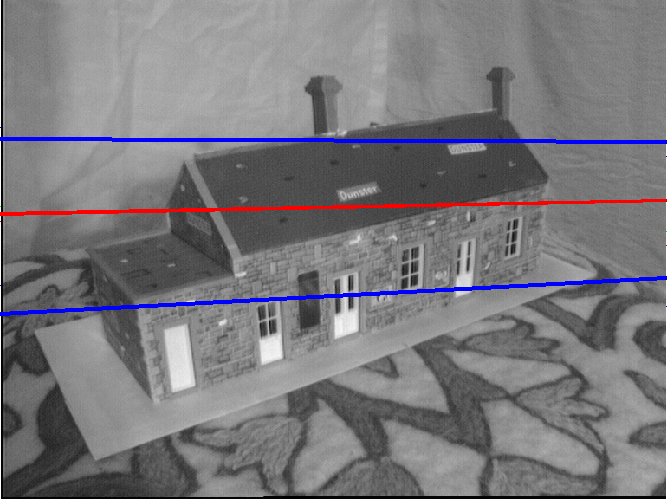}~
b) \includegraphics[width=0.43\linewidth]{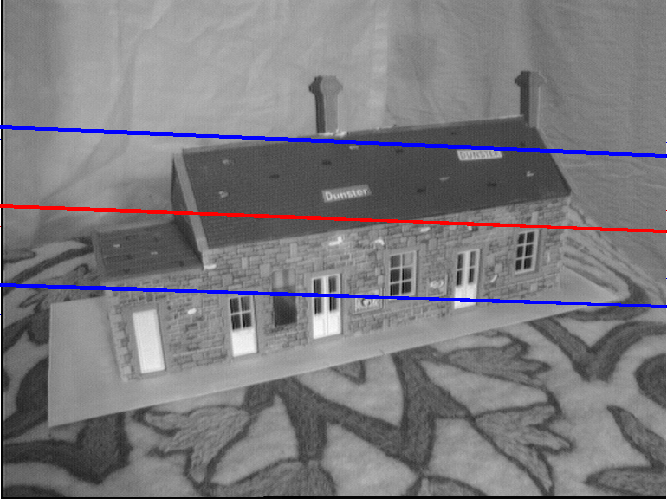}\\~\\
c) \includegraphics[width=0.43\linewidth]{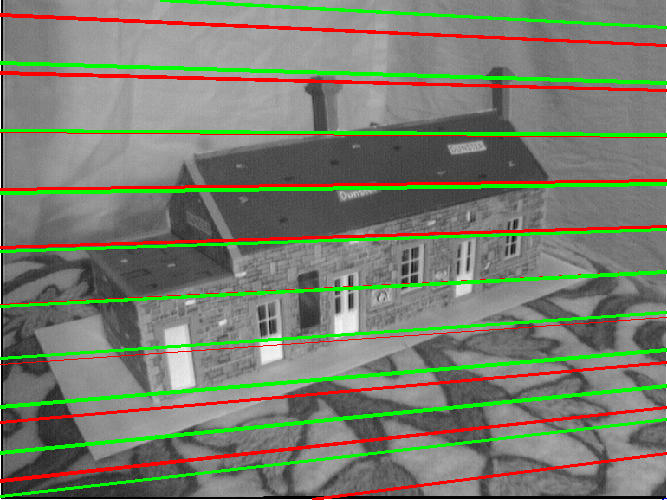}\\
\caption{Initial validation of the epipolar line homography. (a) The first image. The blue lines are the generating lines of the epipole and the red line is the third epipolar line, the angle bisector between them. The best correspondence to the red line is found in Image 2, and from these three lines the homography $H$ from Image 1 to Image 2 is computed. (b) The second image. The red line is the bisector, for which a best correspondence is found in Image 1. This gives an independently estimated homography $G$ from Image 2 to Image 1. (c) The composite homography $GH$ should map epipolar lines to  themselves. The red lines in the first image are transferred forward and  backward using the composite homography. In an ideal case these lines should  overlap.} 
\label{fig:homographyForwardBackward}
\end{figure}

\begin{figure}[tb]
\centering
a) \fbox{\includegraphics[width=0.4\linewidth]{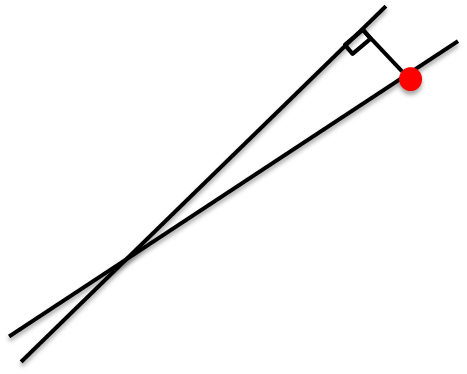}}~
b) \fbox{\includegraphics[width=0.4\linewidth]{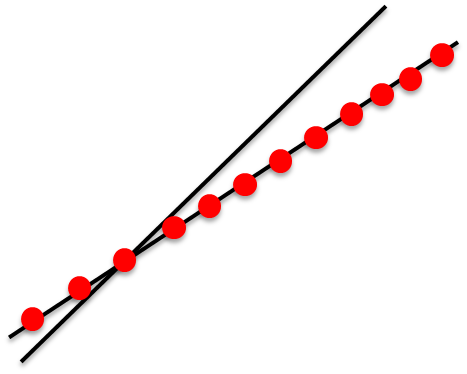}}\\~\\
c) \includegraphics[width=0.43\linewidth]{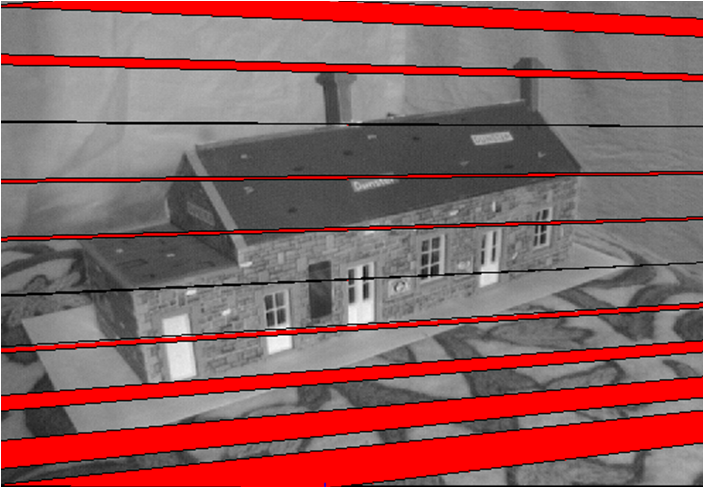}
\caption{The epipolar area measure. (a) The epipolar distance of one point. (b) Considering all the point on the line and evaluating their symmetric epipolar distance is the area between the two lines. (c) The total epipolar area is evaluated for several sampled epipolar lines.}
\label{fig:areaEpipolar}
\end{figure}

\begin{enumerate}[leftmargin=*]

\item Through each pair of the selected points we generate a set of pairs of epipolar line candidates $\{l^i_{p_1}, l^i_{p_2} \}$.
$l^i_{p_1}$ and $l^i_{p_2}$ will be considered candidates only if the second ($l^i_{p_2}$) is closest to the first ($l^i_{p_1}$) and the first is closest to the second (mutual best matches), using the distance of Eq.~\ref{eq:disparity}. This step creates two sets of pairs of lines, one set for $(p_1, p_2)$ and another set for $(q_1, q_2)$.
See Fig.~\ref{fig:overview}.a.

\item Iterate: 

\begin{enumerate}[leftmargin=*]

\item A candidate pair of epipolar lines is sampled from each 2 point pair set generated in Step 2, see Fig.~\ref{fig:overview}.b. Their intersections in each image, 
$e_1=l^i_{p_1} \times l^j_{q_1}$ in Image 1, and 
$e_2=l^i_{p_2} \times l^j_{q_2}$ in Image 2, 
are the hypothesized epipoles, see Fig.~\ref{fig:overview}.c.

\item A third corresponding pair of epipolar lines, $\{l_{e_1}, l_{e_2}\}$, is found. The line
passing through the epipole $e_1$ in image $I_1$ is taken as the bisector of the two lines that generated the epipole. The corresponding line, passing through the epipole $e_2$ in image $I_2$, is found by searching the closest line, in terms of stereo distance, from the pencil of lines passing through the epipole in Image 2.
This is shown in Fig.~\ref{fig:overview}.d.
The epipolar line homography $H$ from Image 1 to Image 2 is computed from the three pairs of corresponding epipolar lines, 
$\{l^i_{p_1}, l^i_{p_2}\}$,
$\{l^j_{q_1}, l^j_{q_2}\}$, and
$\{l_{e_1}, l_{e_2}\}$.

\item Another corresponding pair of epipolar lines is found, but this time a bisector epipolar line is selected in Image 2, and a corresponding epipolar lines is searched for in the pencil of lines passing through the epipole in Image 1. The epipolar line homography $G$ from Image 2 to Image 1 is now computed. See Fig.~\ref{fig:homographyForwardBackward}.

\item The above epipolar line homographies will be correct only if $H=G^{-1}$, and each epipolar line in Image 1 should satisfy $\{l^i \approx G H l^i\}$. This can be measured by computing the area between $l_i$ and $G H l_i$ as in Fig. \ref{fig:areaEpipolar}.

\end{enumerate}
\item 
From all homographies computed in the previous step, we further examine  the 5\% having the highest validation score. For these high-ranked homographies we perform a full validation process.
\item
Full  Validation: 
We sample a large number of epipolar lines through the epipole in one image and transfer them to the other image using the epipolar line homography. For each pair of epipolar lines we measure the line distance according to Eq.~\ref{eq:disparity} and sum over all lines. 
\item
Using the best homography $H$ from the previous step, we compute the fundamental matrix:
$F=[e_2]_\times H^{-1}$. 
\end{enumerate}

\subsection{Remarks}
\begin{itemize}[leftmargin=*]
\item
Robustness is improved by increasing the number of candidates. Instead of examining only pairs of lines that are mutual best matches, we only require them to be in the top 2 matches of each other.
\item
If 3 (instead of 2) corresponding pairs of point matches are given, the algorithm can be significantly accelerated. 
\item
The stereo matching is computed with an $O(N)$ time dynamic program.
\end{itemize}

\subsubsection{Complexity Analysis} 

Let the image sizes be $N \!\times\! N$.
Through each point we sample $O(N)$ lines.
We compare, using stereo matching, $O(N)$,  each sampled line with the $O(N)$ lines sampled at its corresponding point. After $O(N^2)$ comparisons we are left with
$O(N)$ candidate pairs of epipolar lines.
As we are given two points in each image, each point generates $O(N)$ candidate epipolar lines, we get $O(N^2)$ possible intersections of epipolar lines, each such intersection is an hypotheses for the epipoles, $e_1$ and $e_2$. 

For the third pair of corresponding epipolar line we select lines through the epipoles. In $I_1$ we select the bisector of the two epipolar lines that generated $e_1$. We find the best match for this line in $I_2$ by comparing it to $O(N)$ lines through $e_2$. 
This step can be skipped if there is a third pair of corresponding points $r_1$ and $r_2$, then 
$r_1 \times e_1$ and $r_2 \times e_2$ is the third epipolar line pair.

Validation is carried out by taking the epipolar area measure (See Fig.~\ref{fig:areaEpipolar}) of a fixed number of epipolar line in $I_1$ through $e_1$, $\{ l^i_{e1}, GH  l^i_{e1}\}$ with a complexity of $O(1)$.

For two points finding the possible pairs of epipoles together with finding the  third line pair and validation takes $O(N^4)$ steps.
When the algorithm is based on 3 pairs of corresponding points it requires at most $O(N^3)$ steps.

In practice, after filtering lines with little texture, a much smaller number of iterations is required. 

\section{Experiments}
\label{sec:Experiments}

\begin{figure}[tb]
\centering
\includegraphics[width=0.43\linewidth]{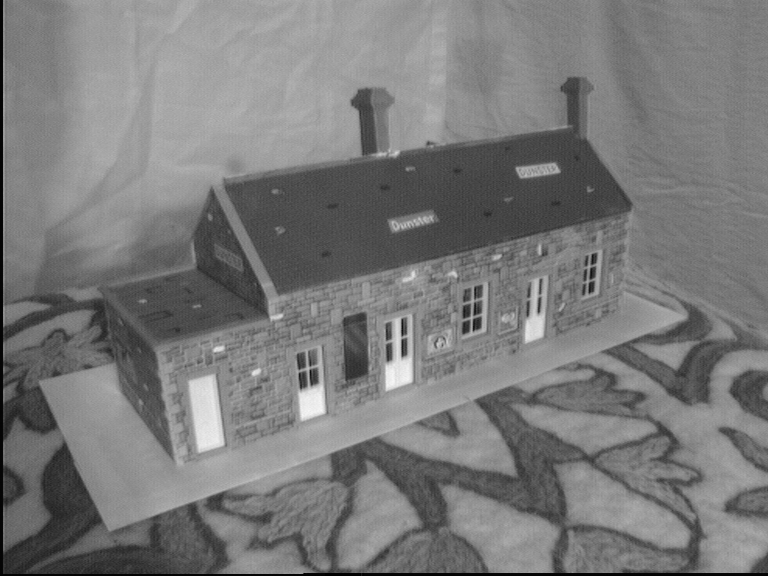}~
\includegraphics[width=0.43\linewidth]{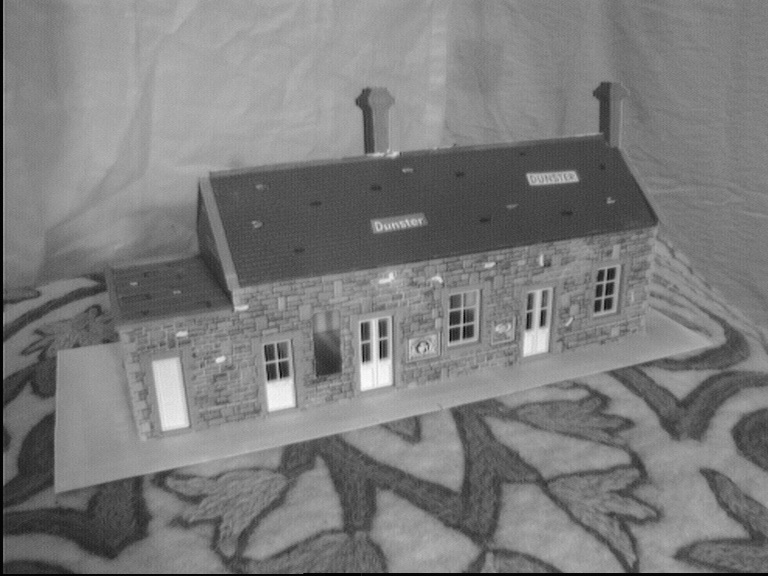}\\~\\
\includegraphics[width=0.43\linewidth]{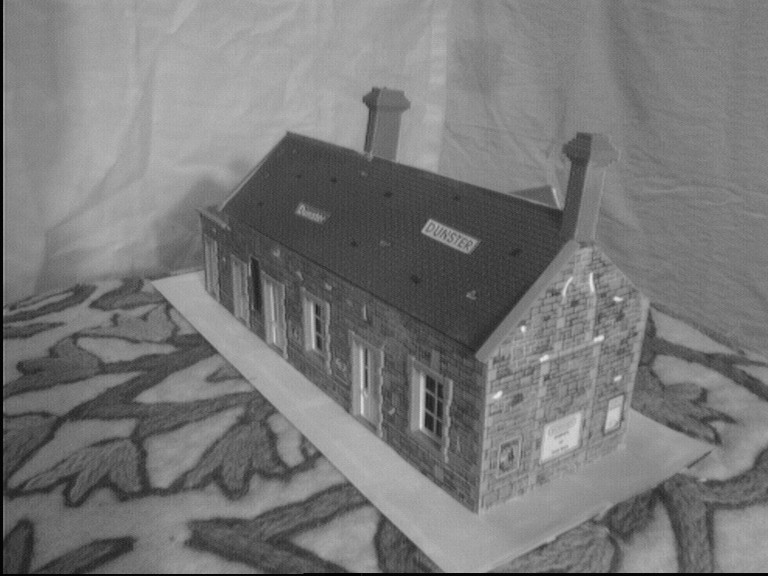}~
\includegraphics[width=0.43\linewidth]{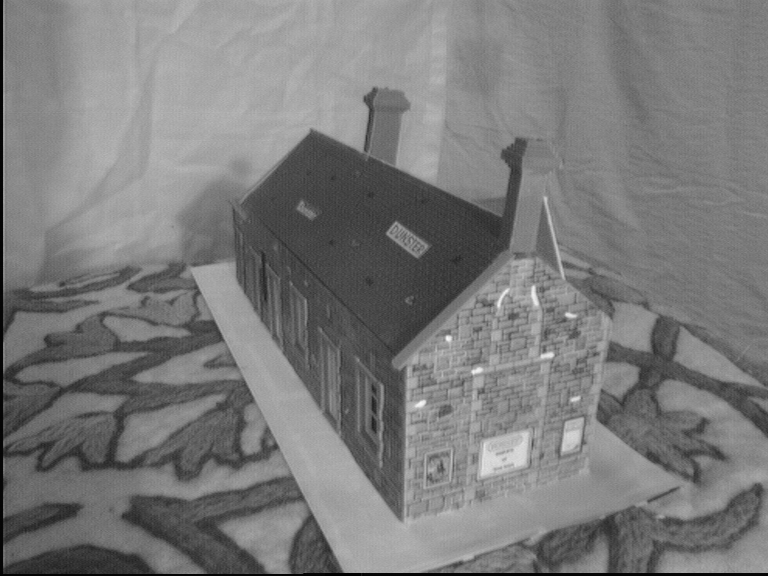}
\caption{The house dataset from VGG is used for real data experiments. The dataset includes 10 images with various angles. Ground truth points and camera matrices are available.}
\label{fig:dataset}
\end{figure}

We evaluated our method using stereo images of the house dataset by VGG, Oxford university \cite{vgg}. The house dataset includes 10 images, representing different angles. The images are presented in Figure~\ref{fig:dataset}. We used every consecutive pair of images as a stereo pair, which results in 9 pairs.   The size of the images is $768 \times 576$. 

The quality of the computed fundamental matrices was evaluated using the symmetric epipolar distance \cite{hartley2003multiple} with respect to the given ground truth points. The baseline method is the  $7$-point algorithm. The $7$-point method returns 3 possible solutions. In all experiments we selected for comparison the solution with the lowest symmetric epipolar distance. We also compared with the $8$-point algorithms, which returned high-quality solutions after data normalization. Both the 7-point and 8-point algorithm were computed using the VGG toolbox\cite{vggtoolbox}. 

For each pair of images we repeatedly executed 10 iterations. In each iteration we randomly sampled two pairs of corresponding points as input to our approach, seven pairs of corresponding points to use as input to the $7$-point method and eight pairs of corresponding points to use as input to the $8$-point method. The points were sampled so that they  were at least $30$ pixels apart to ensure stability.
We computed the fundamental matrix using our method,  the $7$-point method, and the $8$-point method. The symmetric epipolar distance was computed for each method.  
The points were sampled so that they  were at least $30$ pixels apart to ensure stability.

\begin{figure}[tb]
\begin{center}
\begin{tabular}{|c|c|c|c|c|c|}
\hline
	  \multicolumn{3}{|c|}{Image Pairs} &  7-point & 8-point & 2-point\\ \hline       
      1 & \includegraphics[height=0.8cm]{images/house000.png} & \includegraphics[height=0.8cm]{images/house001.png} & 27.11 & 3.27 & 2.54 \\ \hline
      2 & \includegraphics[height=0.8cm]{images/house001.png}& \includegraphics[height=0.8cm]{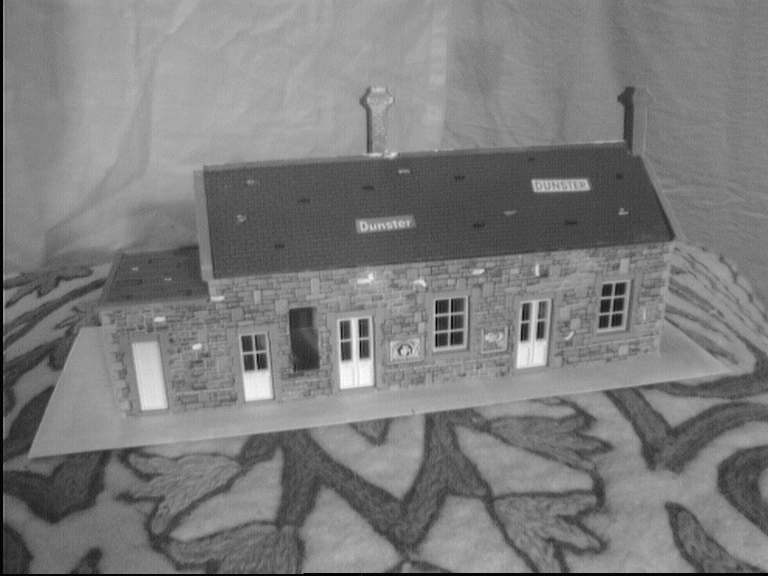} & 25.80 & 2.11 & 2.91   \\ \hline
      3 & \includegraphics[height=0.8cm]{images/house002.png}& \includegraphics[height=0.8cm]{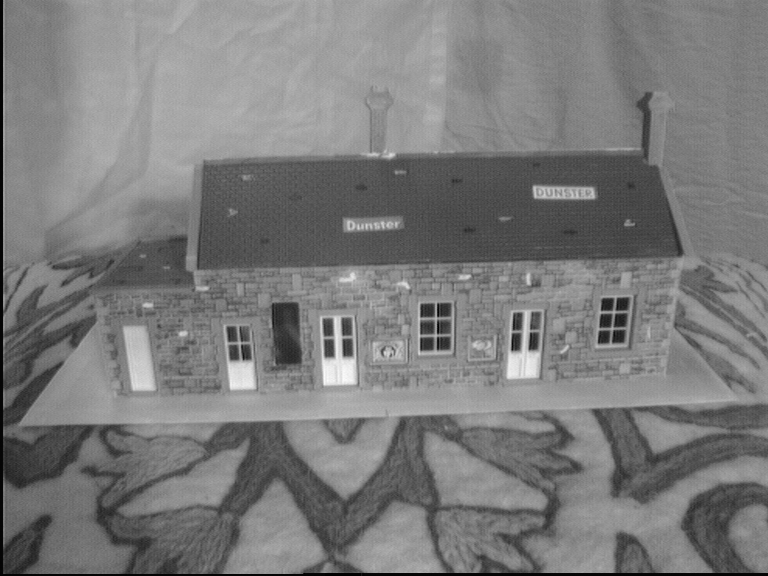} & 27.14& 2.02 & 2.01    \\  \hline    
      4 & \includegraphics[height=0.8cm]{images/house003.png}& \includegraphics[height=0.8cm]{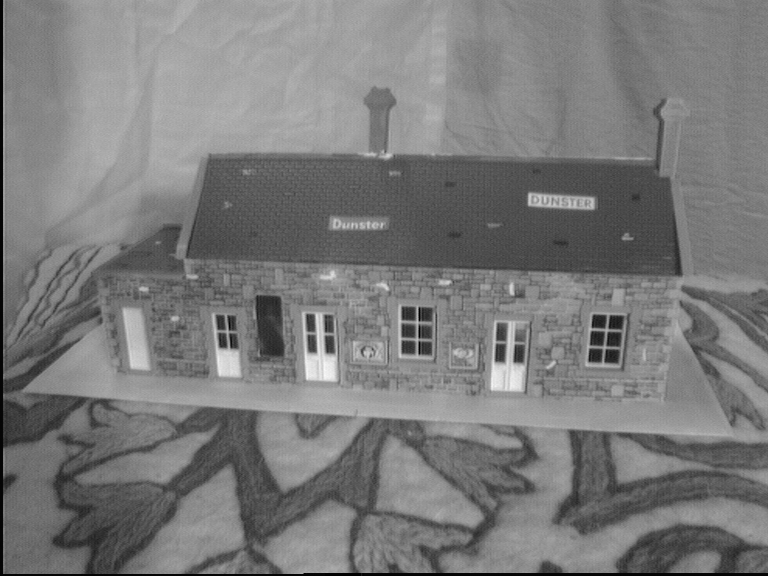} & 27.11 & 2.5 & 2.04   \\ \hline       
      5 & \includegraphics[height=0.8cm]{images/house004.png}& \includegraphics[height=0.8cm]{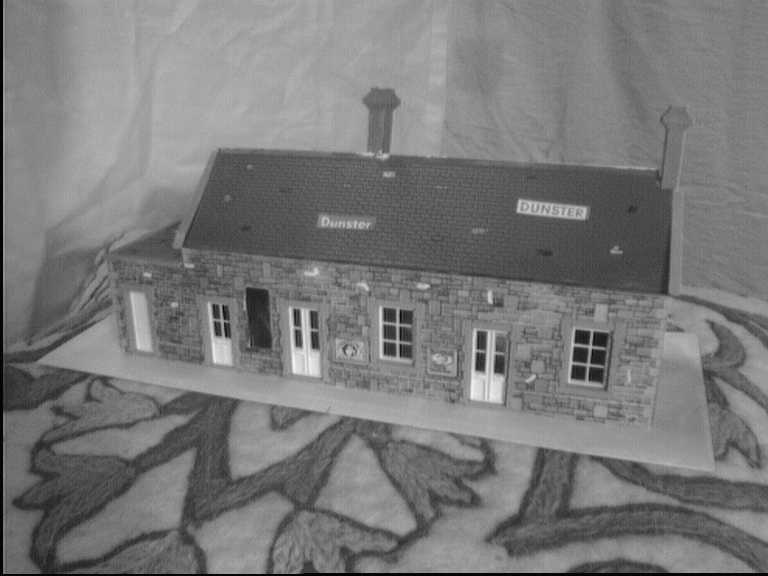} & 25.47 & 2.00 & 2.34  \\ \hline
      6 & \includegraphics[height=0.8cm]{images/house005.png} & \includegraphics[height=0.8cm]{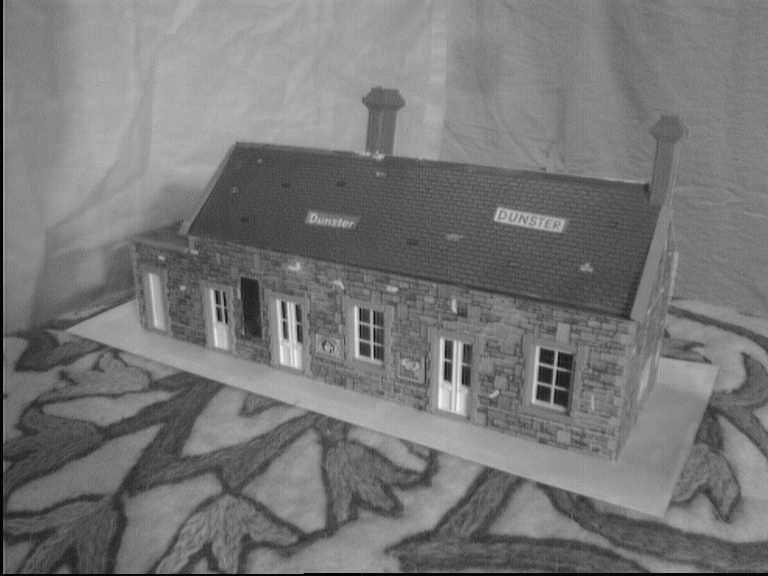} & 14.67& 2.77 & 4.12 \\ \hline      
      7 & \includegraphics[height=0.8cm]{images/house006.png}& \includegraphics[height=0.8cm]{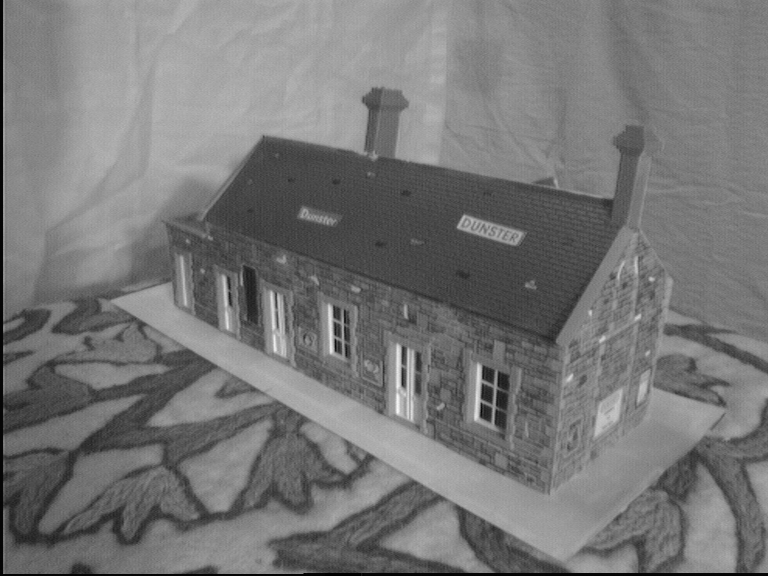} & 18.87& 3.7 & 5.76   \\ \hline      
      8 & \includegraphics[height=0.8cm]{images/house007.png} & \includegraphics[height=0.8cm]{images/house008.png} & 18.97& 4.22 & 5.66   \\ \hline      
	 9 & \includegraphics[height=0.8cm]{images/house008.png}& \includegraphics[height=0.8cm]{images/house009.png} & 26.13 & 5.89 & 2.30   \\ \hline      
\end{tabular}
\end{center}
\caption{
The symmetric epipolar distance of the estimated fundamental matrix using the 7-point/8-point  algorithm, and our 2-point algorithm. The distance is with respect to ground truth points. Accuracy of our 2-point algorithm is substantially higher than the $7$-point algorithm and slightly better than the $8$-point algorithm. The median error of our algorithm is $2.54$. For the $8$-point algorithm the median is $2.77$  while for the 7-point it is $25.8$.
\label{table:results1}}
\end{figure}

\begin{figure}[tb]
\centering
\includegraphics[width=0.99\linewidth]{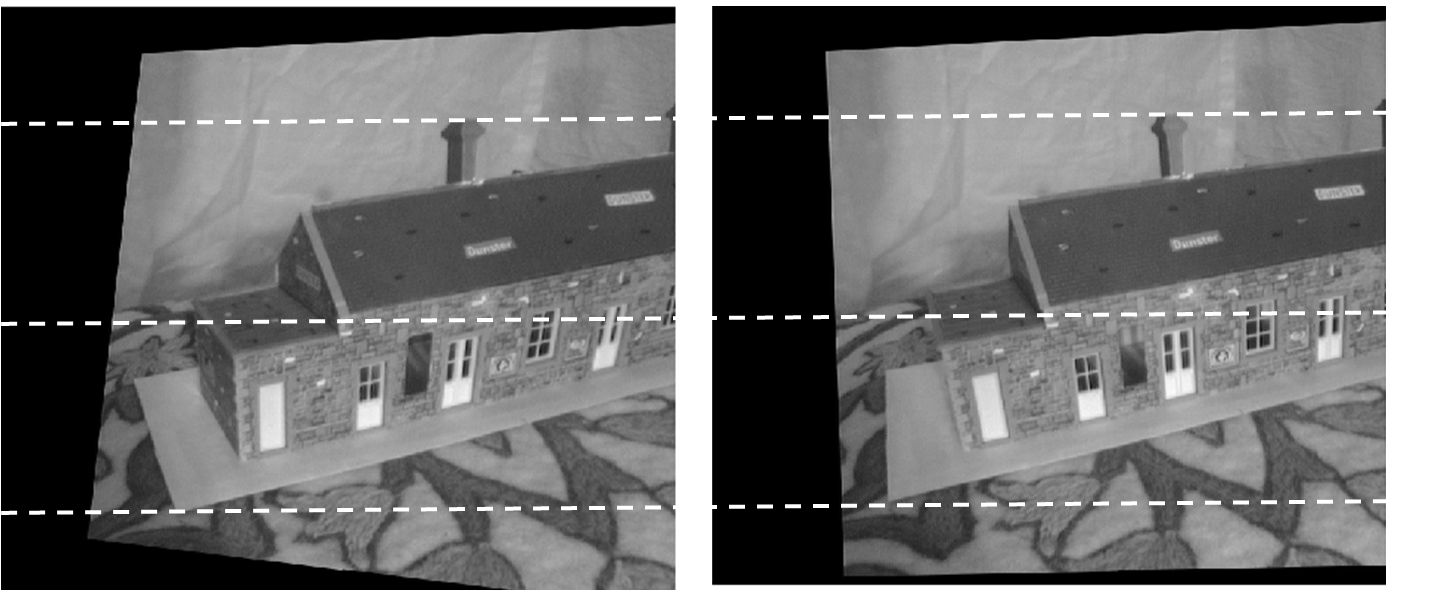}
\caption{Rectification example, the images are rectified after estimation of the fundamental matrix using our approach from only two pairs of corresponding points. Each horizontal line in the images is a pair of epipolar lines. Corresponding feature points are  on the same epipolar lines.}
\label{fig:preview}
\end{figure}

Fig.~\ref{table:results1} shows the resulting fundamental matrix for each pair of images. Our method significantly outperforms the the $7$-point algorithm. In $66\%$ of the cases the symmetric epipolar distance is less than 3 pixels. The median error in our approach is $2.54$.  For the $8$-point algorithm the median is $2.77$  while the median in the 7-point algorithm is $25.8$. Our approach depends on  \emph{global} intensity matching rather than on the exact matching of the points. Pairs $7,8$ introduce a challenging stereo matching and as a result the quality of our method is  effected. The global intensities in pairs 1-6 can be accurately matched and as a result the quality of the estimated fundamental matrix is high. Fig.~\ref{fig:preview} shows an example of rectification using our estimated fundamental matrix for pair number 1. Each horizontal line in the images is a pair of epipolar lines. Corresponding feature points are placed on the same epipolar lines. 

\section{Conclusions}

We presented a method to compare lines, based on stereo matching, suitable for finding corresponding epipolar lines. This can be used to compute the fundamental matrix based on only three such line correspondences.

Finding corresponding epipolar lines is greatly accelerated if we have 2 matching points. In this case our algorithm is  very robust and fast with accuracy greatly outperforming  the 7 point method and competitive with 8 point methods.

~\\
\noindent
{\bf Acknowledgment.} This research was supported by Google, by Intel ICRI-CI, by DFG, and by the Israel Science Foundation.

\bibliographystyle{plain}
\bibliography{egbib}

\end{document}